\documentclass{article}

\PassOptionsToPackage{numbers, compress}{natbib}

    \usepackage[preprint]{neurips_2024}

\usepackage{amsmath}
\usepackage{amssymb}
\usepackage{amsthm}

\usepackage[utf8]{inputenc} 
\usepackage[T1]{fontenc}    
\usepackage{hyperref}       
\usepackage{url}            
\usepackage{booktabs}       
\usepackage{amsfonts}       
\usepackage{nicefrac}       
\usepackage{microtype}      
\usepackage{xcolor}         

\usepackage{bm}

\usepackage{subcaption}

\usepackage{multirow} 
\usepackage{makecell}

\usepackage{xcolor}
\usepackage{tikz}

\definecolor{myred}{rgb}{1.0, 0.62, 0.61}
\definecolor{mygreen}{RGB}{8, 147, 146}
\definecolor{myorange}{RGB}{233, 159, 105}
\definecolor{mypink}{RGB}{207, 89, 126}

\usepackage{setspace}

\AtEndPreamble{
    \usepackage[capitalize]{cleveref}
    \crefname{section}{Sec.}{Secs.}
    \Crefname{section}{Section}{Sections}
    \Crefname{table}{Table}{Tables}
    \crefname{table}{Tab.}{Tabs.}
    \Crefname{proposition}{Proposition}{Propositions}
    \crefname{proposition}{Prop.}{Props.}
    \Crefname{subproposition}{Part}{Parts}
    \crefname{subproposition}{Part.}{Parts.}
}
\usepackage{enumitem}

\makeatletter
\renewcommand
\tableofcontents
{
  \begin{spacing}{1.0}
  \large
  \@starttoc{toc}%
  \end{spacing}
}
\makeatother

\usepackage{wrapfig}

\usepackage{fontawesome}

\begin{document}
\addtocontents{toc}{\protect\setcounter{tocdepth}{-1}}
\title{KAN or MLP: A Fairer Comparison}

\author{
  Runpeng Yu, Weihao Yu, and Xinchao Wang\\
  National University of Singapore \\
}

\maketitle

\vspace{-2em}
\begin{center}
\faGithub \, \href{https://github.com/yu-rp/KANbeFair}{https://github.com/yu-rp/KANbeFair}
\end{center}

\vspace{-1em}
\begin{center}
     \begin{minipage}{0.95\textwidth}\centering
     \textbf{TL;DR-}\textit{Under the same number of parameters or FLOPs, we find KAN  outperforms MLP only in symbolic formula representing, but remains inferior to MLP on other tasks of machine learning, computer vision, NLP, and audio processing.
     }
     \end{minipage}
\end{center}

\begin{figure}[htbp]
    \centering
    \includegraphics[width=\textwidth]{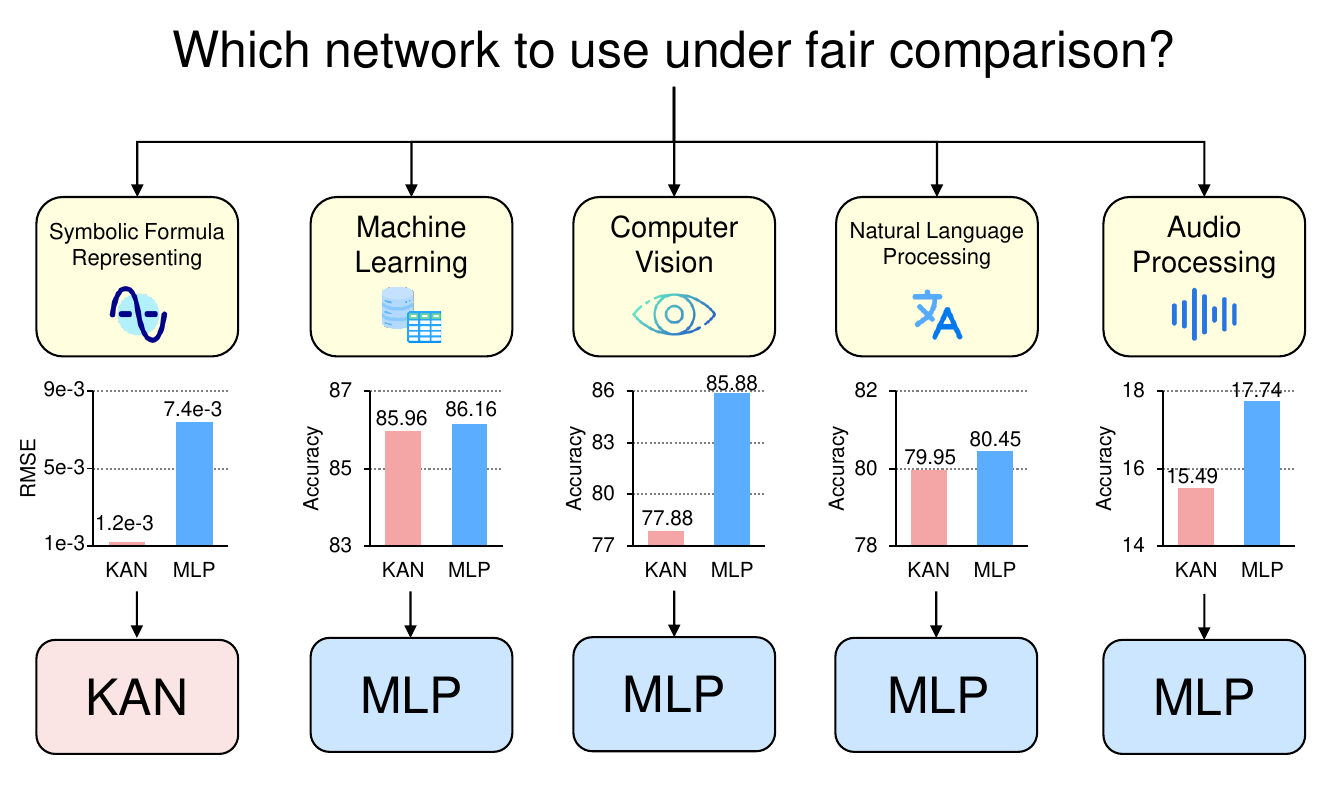}
    \caption{
    Performance comparison between KAN and MLP under fair setup.
    MLP yields higher average accuracy 
    in machine learning, computer vision, natural language processing, and audio processing,  while KAN leads to lower average root mean square error.
    For the Symbolic Formula Representation task, a lower RMSE is better.
    }
    \label{fig:images}
\end{figure}

\begin{abstract}

This paper does not introduce a novel method. Instead, it offers a fairer and more comprehensive comparison of KAN and MLP models across various tasks, including machine learning, computer vision, audio processing, natural language processing, and symbolic formula representation. Specifically, we control the number of parameters and FLOPs to compare the performance of KAN and MLP. Our main observation is that, except for symbolic formula representation tasks, MLP generally outperforms KAN. We also conduct ablation studies on KAN and find that its advantage in symbolic formula representation mainly stems from its B-spline activation function. 
When B-spline is applied to MLP, performance in symbolic formula representation significantly improves, surpassing or matching that of KAN. However, in other tasks where MLP already excels over KAN, B-spline does not substantially enhance MLP’s performance. Furthermore, we find that KAN's forgetting issue is more severe than that of MLP in a standard class-incremental continual learning setting, which differs from the findings reported in the KAN paper. We hope these results provide insights for future research on KAN and other MLP alternatives.
\end{abstract}

\section{Introduction}
Multi-Layer Perceptrons (MLP) \cite{mcculloch1943logical, rosenblatt1958perceptron,rumelhart1986learning}, also known as fully-connected feedforward neural networks, are fundamental building blocks in modern deep learning models. MLP consist of multiple layers of nodes, where each node (or neuron) applies a fixed activation function to a weighted sum of its inputs. This structure allows MLP to approximate a wide range of nonlinear functions, a property guaranteed by the universal approximation theorem. As a result, MLP are extensively used in various deep learning tasks, including classification, regression, and feature extraction. Despite their versatility and widespread application, MLP have certain limitations, such as difficulty in interpreting the learned representations and flexibility in expanding the network scale.

Kolmogorov–Arnold Networks (KANs)~\cite{liu2024kan} presents an innovative alternative to traditional MLP by leveraging the Kolmogorov-Arnold representation theorem. Unlike MLP, which employ fixed activation functions on nodes, KANs feature learnable activation functions on edges, replacing linear weight parameters with univariate functions parameterized as splines. KANs are expected to be promising alternatives for MLP \cite{liu2024kan}, which motivated us to carefully examine it. However, current comparison experiments between KAN and MLP are not fair, under different parameters or FLOPs. To probe the potential of KAN, it is necessary to comprehensively compare KAN and MLP under fair settings.

To this end, we control the parameters or FLOPs of KAN and MLP,  and then train and evaluate them on tasks in various domains, including  symbolic formula representing, machine learning, computer vision, NLP, and audio processing. Under these fair settings, we observe that KAN outperforms MLP only in symbolic formula representation tasks, while MLP typically excels over KAN in other tasks.

We further find that KAN's advantages in symbolic formula representation stem from its use of the B-spline activation function. Originally, MLP's overall performance lags behind KAN, but after replacing MLP's activation function with B-spline, its performance matches or even surpasses that of KAN. However, BSpine cannot further improve the performance of MLP on other tasks, such as computer vision. 

We also found that KAN does not actually perform better than MLP in continual learning task. The original KAN paper compared the performance of KAN and MLP in continual learning tasks using a series of one-dimensional functions, where each subsequent function was a translation of the previous function along the number axis. We compared the performance of KAN and MLP in a more standard class-incremental continual learning setting. Under fixed training iterations, we found that KAN's forgetting issue is more severe than that of MLP. We hope that these results can provide insights for future research on KAN and other MLP alternatives.

\section{Formulation of KAN and MLP}
There are two branches in KAN, the first branch is the B-spline branch, and the other branch is the shortcut branch, which is a non-linear activation concatenated with a linear transformation. In the official implementation, the shortcut branch is a SiLU function followed by a linear transformation. 
Let $x$ denote the feature vector for one sample. Then, the forward equation of KAN`s spline branch can be written as $$x = [\bm{A}*\text{Spline}(\bm{1}x^T)]\bm{1}.$$ Here the Spline function is an element-wise non-linear function. Its input is a matrix and its output is the matrix with the same shape as the input. For each element in the matrix, the spline function has a different non-linear operation. All the non-linear operations share a same template, but with different parameters. In the original KAN architecture, the spline function is chosen to be the B-spline function. The parameters of each B-spline function are learned together with other network parameters.

Correspondingly, the forward equation of a one-layer MLP with non-linear operation first can be written as 
$$x = [\bm{A}*\sigma(\bm{1}x^T)]\bm{1}.$$
This formula shares the same form as the B-spline branch formula in KAN, differing only in the nonlinear function. Therefore, aside from the interpretation of the KAN structure in the original KAN paper, KAN can also be viewed as a type of fully-connected layer. This fully-connected layer undergoes an elementwise nonlinear function, followed by a linear transformation. The differences between KAN and common MLP are in two aspects. (1) Different activation functions. 
Commonly, the activation functions in MLP, such as ReLU and GELU, have no learnable parameters and are uniform for all input elements.  However, in KAN, the activation function is a spline function, which has learnable parameters and is different for each input element. (2) The order of linear and non-linear operation. Generally, we conceptualize MLP as performing a linear transformation followed by a non-linear transformation. However, KAN actually performs a non-linear transformation first, followed by a linear transformation. 
To some extent, it is also feasible to describe the fully-connected layers in an MLP as ``first non-linear, then linear''. For example, for an MLP, the following two descriptions are equivalent: some fully-connected layers with non-linear last and a linear head, or a linear tail and some fully-connected layers with non-linear first.

Among the differences between KAN and MLP, we believe that the variation in activation functions is the primary factor that differentiates KAN from MLP. We hypothesize that the variation in activation functions makes KAN and MLP suitable for different tasks, resulting in functional differences between the two models. To validate this hypothesis, we compared the performance of KAN and MLP across various tasks and delineated the tasks each model suits. To ensure a fair comparison, we first derived the formulas for calculating the number of parameters and FLOPs for KAN and MLP. In our experiments, we controlled for either the same number of parameters or FLOPs to compare the performance of KAN and MLP.

\section{Number of Parameters of KAN and MLP}
We use $d_{in}$ and $d_{out}$ to denote the input and output dimensions of the neural network layer.
We use $K$ to denote the order of the spline, which corresponds to the parameter \textit{k} of the official nn.Module KANLayer. It is the order of the Polynomial basis in the spline function. We use $G$ to denote the number of spline intervals, which corresponds the \textit{num} parameter of the official nn.Module KANLayer. It is the number of intervals of the B-spline, before padding. It equals the number of control points - 1, before padding. After padding, there should be $(K + G)$ functioning control points.

The learnable parameters in KAN include the B-spline control points, the shortcut weights, the B-spline weights, and the bias. The total learnable parameters are 
$${(d_{in}\times d_{out})\times (G+K+3) + d_{out}}.$$

Correspondingly, the learnable parameters of one MLP layer is 
$${(d_{in}\times d_{out}) + d_{out}}.$$

\section{FLOPs of KAN and MLP}
In our evaluation, the FLOPs for any arithmetic operations are considered to be 1, and the FLOPs for Boolean operations are considered to be 0. The operations in De Boor-Cox algorithm at order 0 can be transformed into a series of Boolean operations where it does not require floating-point operations. Thus, theoretically, its FLOPs is 0. This is different from the official KAN implementation, in which it transforms the Boolean data back to Float data to conduct the operation. In our evaluation, the FLOPs is calculated for one sample. The B-spline FLOPs with De Boor-Cox Iterative Formulation implementation in the official KAN code is 
    \begin{align*}
        (d_{in}\times d_{out}) \times &[ 9\times K\times(G+1.5\times K) + 2\times G-2.5\times k-1].
    \end{align*}
Together with the FLOPs of the shortcut path and the FLOPs of merging the two branches, the total FLOPs of one KAN layer is  
    \begin{align*}
        &\text{FLOPs of non-linear function} \times d_{in} \\
        &\quad+(d_{in}\times d_{out}) \times [ 9\times K\times(G+1.5\times K) + 2\times G-2.5\times K+3].
    \end{align*}

Correspondingly, the FLOPs of one MLP layer is 
$$2\times{d_{in}\times d_{out} + \text{FLOPs of non-linear function} \times d_{out}}.$$

The difference between the FLOPs of a KAN layer and that of a MLP layer with the same input dimension and output dimension can be written as 
    \begin{align*}
        &\text{FLOPs of non-linear function} \times (d_{in} - d_{out}) \\
        &\quad+(d_{in}\times d_{out}) \times [ 9\times K\times(G+1.5\times K) + 2\times G-2.5\times K+1].
    \end{align*}
If the MLP also does the non-linear operation first, the first term is zero.

\section{Experiments}

Our objective is to compare the performance differences between KAN and MLP under the same number of parameters or FLOPs. Our experiments cover various tasks including machine learning, computer vision, natural language processing, audio processing, and symbolic formula representing. All experiments used the Adam optimizer with a batch size of 128, and learning rates of 1e-3 or 1e-4. All experiments are conducted on one RTX3090 GPU. 

\subsection{Performance Comparison}

\begin{figure}[h!]
  \centering
  \includegraphics[width=0.9\textwidth]{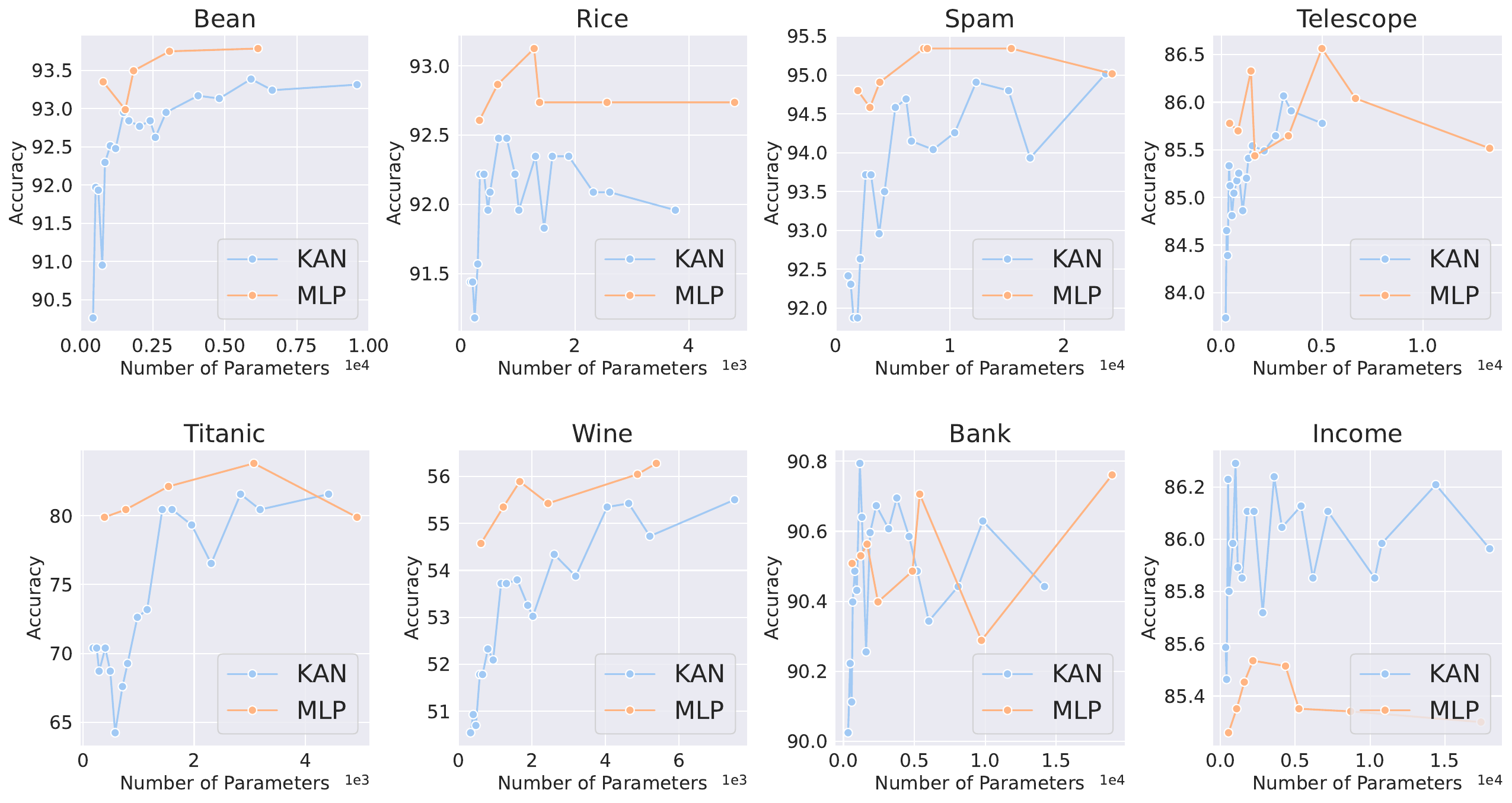}
  \caption{Performance comparison between KAN and MLP on Machine Learning datasets controlling the number of parameters.}
  \label{fig:ml_para}
\end{figure}

\begin{figure}[h!]
  \centering
  \includegraphics[width=0.9\textwidth]{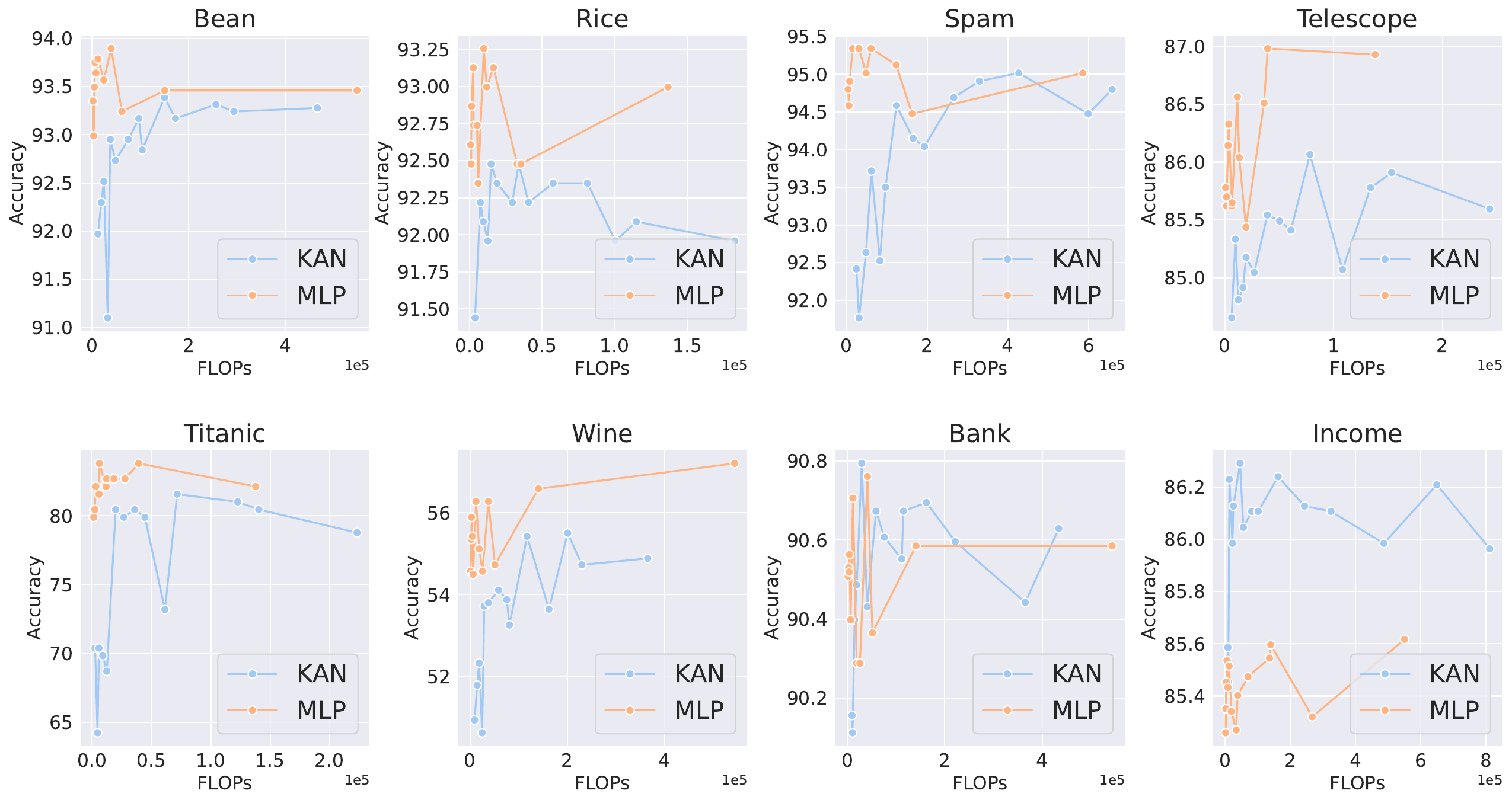}
  \caption{Performance comparison between KAN and MLP on Machine Learning datasets controlling FLOPs.}
  \label{fig:ml_flops}
\end{figure}

\textbf{Machine Learning.} We conduct experiments on 8 machine learning datasets. We use KAN and MLP with one or two hidden layers. The input and output dimensions of the neural networks were adjusted according to the datasets. The hidden layer widths for MLP were 32, 64, 128, 256, 512, or 1024, with GELU or ReLU as the activation functions, and normalization layers were used in the MLP. For KAN, the hidden layer widths were 2, 4, 8, or 16, the number of B-spline grids was 3, 5, 10, or 20, and the B-spline degrees were 2, 3, or 5. Since the original KAN architecture does not include normalization layers, to balance the potential advantage from the normalization layers in MLP, we increased the range of KAN's spline function. In addition to the range of [-1,1] used in the original paper, we also experimented with ranges of [-2,2] and [-4,4]. All experiments are trained for 20 epochs. The best accuracy on the testing set during the training is recorded. The upper envelope of all the recorded points is plotted in the \cref{fig:ml_para} and \cref{fig:ml_flops}. On machine learning datasets, MLP generally maintains an advantage. In our experiments on eight datasets, MLP outperformed KAN on six of them. However, we also observed that on one dataset, the performance of MLP and KAN was nearly equivalent, and on another dataset, KAN outperformed MLP. For these two datasets, we will conduct an architecture ablation in the next subsection to analyze the reasons behind KAN's advantages. Overall, MLP still holds a general advantage over KAN on machine learning datasets.

\begin{figure}[h!]
  \centering
  \includegraphics[width=0.9\textwidth]{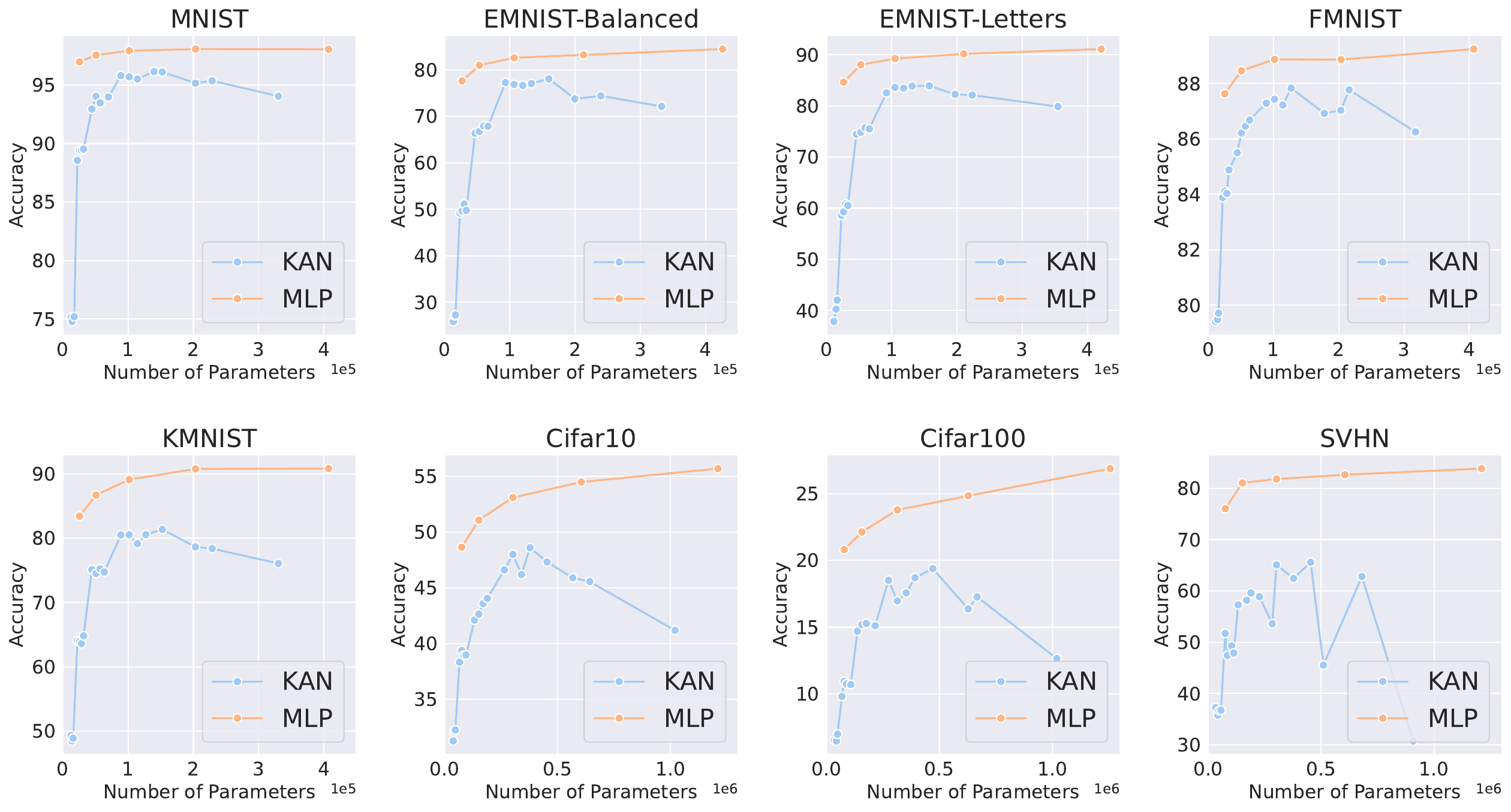}
  \caption{Performance comparison between KAN and MLP on Computer Vision datasets controlling the number of parameters.}
  \label{fig:vision_para}
\end{figure}

\begin{figure}[h!]
  \centering
  \includegraphics[width=0.9\textwidth]{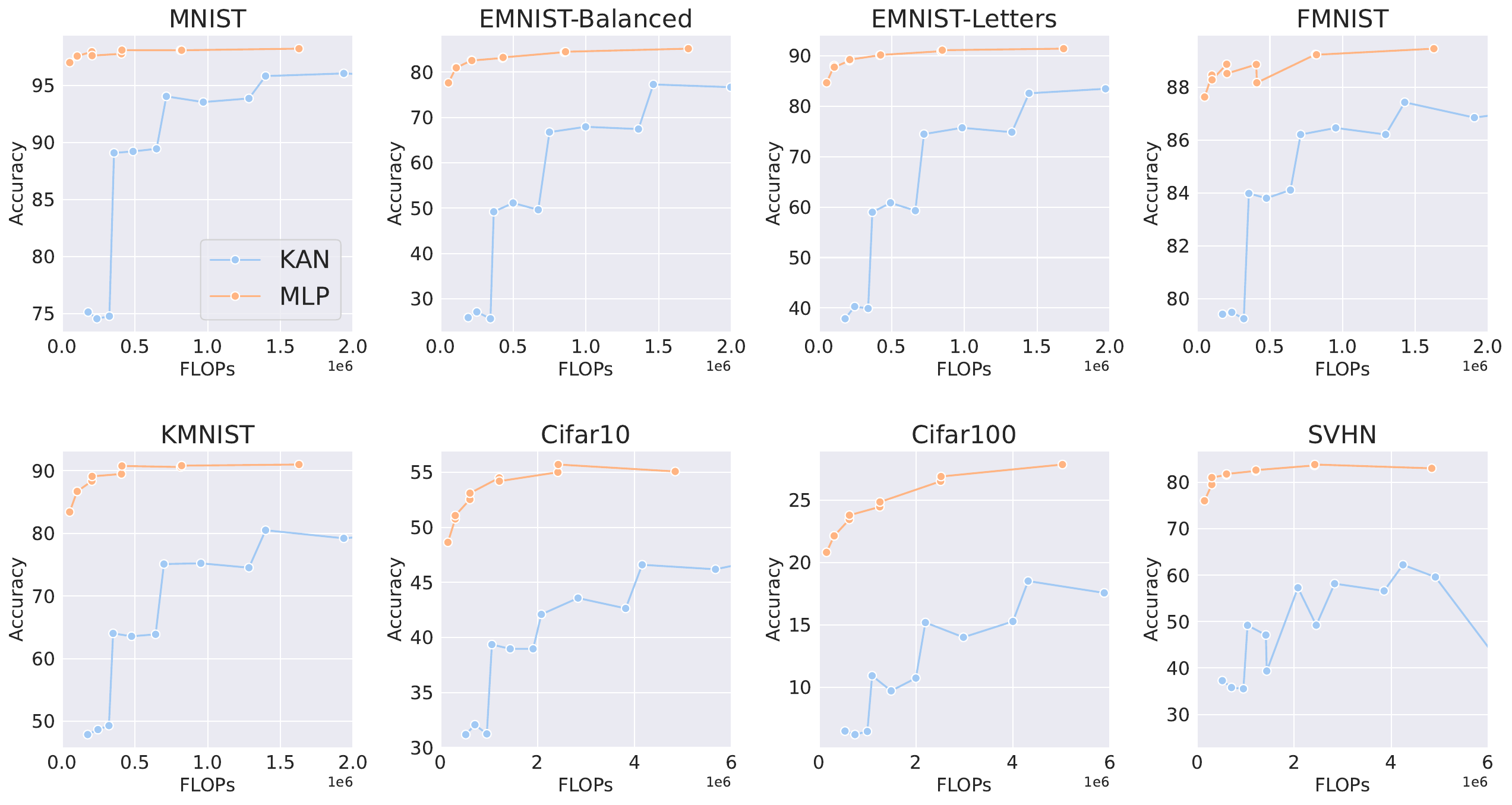}
  \caption{Performance comparison between KAN and MLP on Computer Vision datasets controlling FLOPs.}
  \label{fig:vision_flops}
\end{figure}

\textbf{Computer Vision.} We conduct experiments on 8 computer vision datasets. We use KAN and MLP with one or two hidden layers. The input and output dimensions of the neural networks were adjusted according to the datasets. The hidden layer widths for MLP were 32, 64, 128, 256, 512, or 1024, with GELU or ReLU as the activation functions, and no normalization layers were used. For KAN, the hidden layer widths were 2, 4, 8, or 16, the number of B-spline grids was 3, 5, 10, or 20, and the B-spline degrees were 2, 3, or 5. All experiments are trained for 20 epochs. The best accuracy on the testing set  during the training is recorded. The upper envelope of all the recorded points is plotted in the \cref{fig:vision_para} and \cref{fig:vision_flops}. For the computer vision datasets, the conductive bias introduced by KAN's spline functions was not effective, as its performance consistently fell short of MLP with the same number of parameters or FLOPs.

\begin{figure}[h!]
  \centering
  \includegraphics[width=0.9\textwidth]{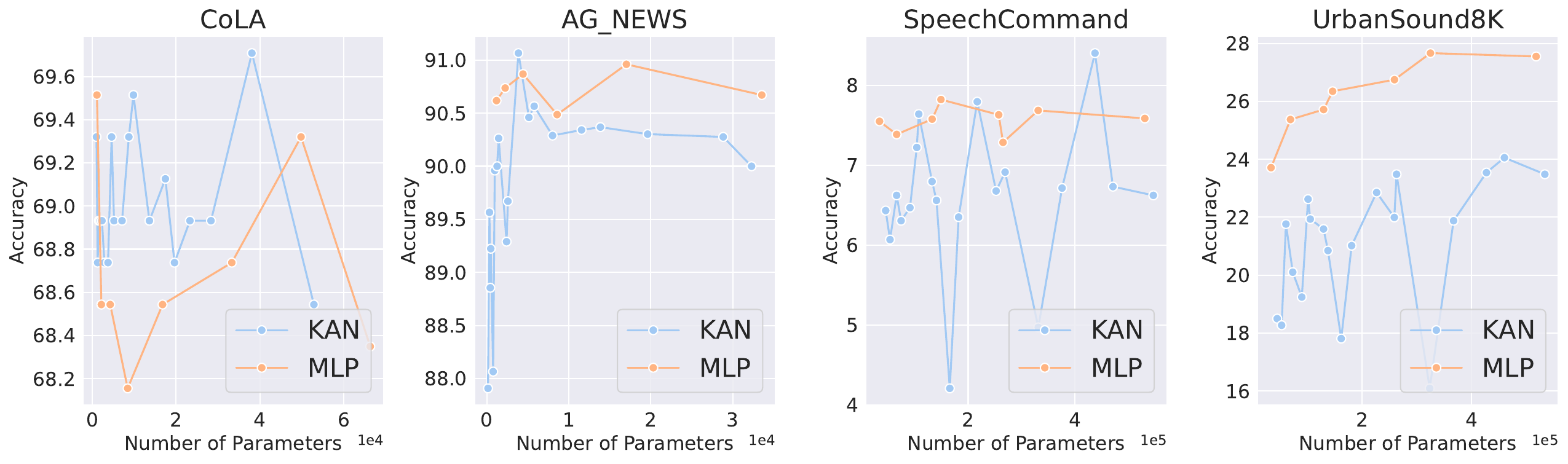}
  \caption{Performance comparison between KAN and MLP on audio and text datasets controlling the number of parameters.}
  \label{fig:audio_text_para}
\end{figure}

\begin{figure}[h!]
  \centering
  \includegraphics[width=0.9\textwidth]{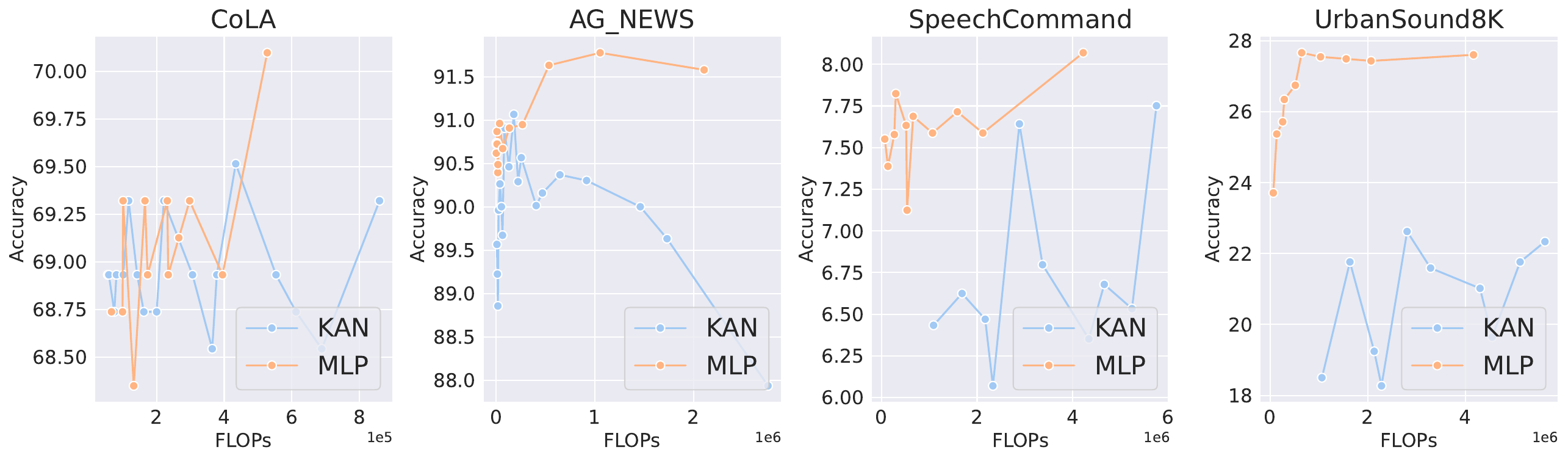}
  \caption{Performance comparison between KAN and MLP on audio and text datasets controlling FLOPs.}
  \label{fig:audio_text_flops}
\end{figure}

\textbf{Audio and Natural Language.} We conduct experiments on 2 audio classification and 2 text classification datasets. We use KAN and MLP with one or two hidden layers. The input and output dimensions of the neural networks were adjusted according to the datasets. The hidden layer widths for MLP were 32, 64, 128, 256, 512, or 1024, with GELU or ReLU as the activation functions, and normalization layers were used in the MLP. For KAN, the hidden layer widths were 2, 4, 8, or 16, the number of B-spline grids was 3, 5, 10, or 20, the B-spline degrees were 2, 3, or 5, and the B-spline range is [-1,1], [-2,2] or [-4,4]. All experiments are trained for 20 epochs, except on the UrbanSound8K dataset, on which the models are trained for 40 epochs. The best accuracy on the testing set during the training is recorded. The upper envelope of all the recorded points is plotted in the \cref{fig:audio_text_para} and \cref{fig:audio_text_flops}. On both two audio datasets, MLP outperformed KAN. In text classification tasks, MLP maintained an advantage on the AG News dataset. However, on the CoLA dataset, there was no significant difference in performance between MLP and KAN. When controlling for the same number of parameters, KAN appeared to have an advantage on the CoLA dataset. Nevertheless, due to the high FLOPs required by KAN's spline functions, this advantage did not persist in experiments controlling for FLOPs. When FLOPs were controlled, MLP seemed to be superior. Thus, there is no consistent answer as to which is better on the CoLA dataset. Overall, MLP remains a better choice for audio and text tasks.

\begin{figure}[h!]
  \centering
  \includegraphics[width=0.9\textwidth]{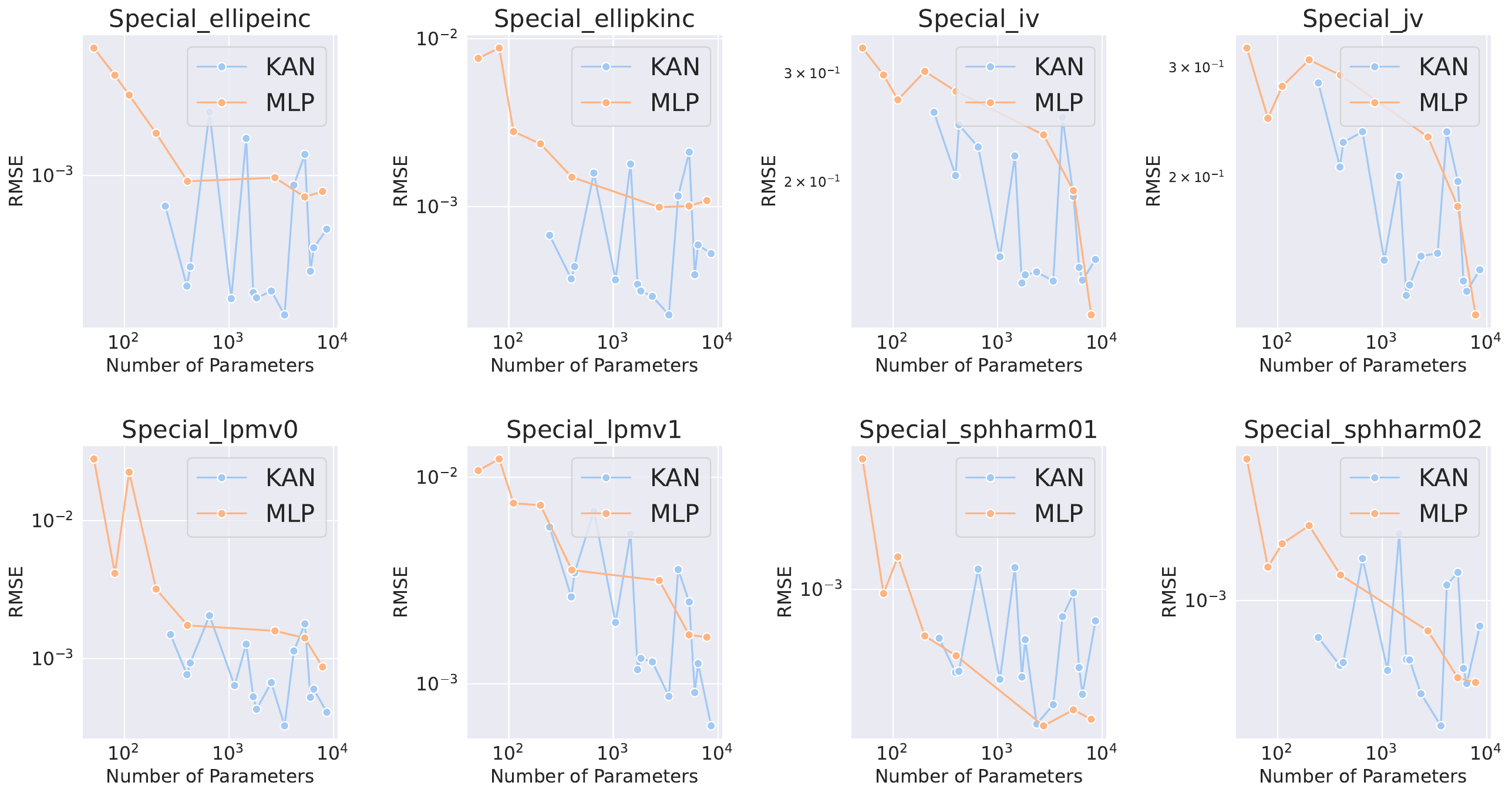}
  \caption{Performance comparison between KAN and MLP for symbolic formula representing controlling the number of parameters.}
  \label{fig:special_para}
\end{figure}

\begin{figure}[h!]
  \centering
  \includegraphics[width=0.9\textwidth]{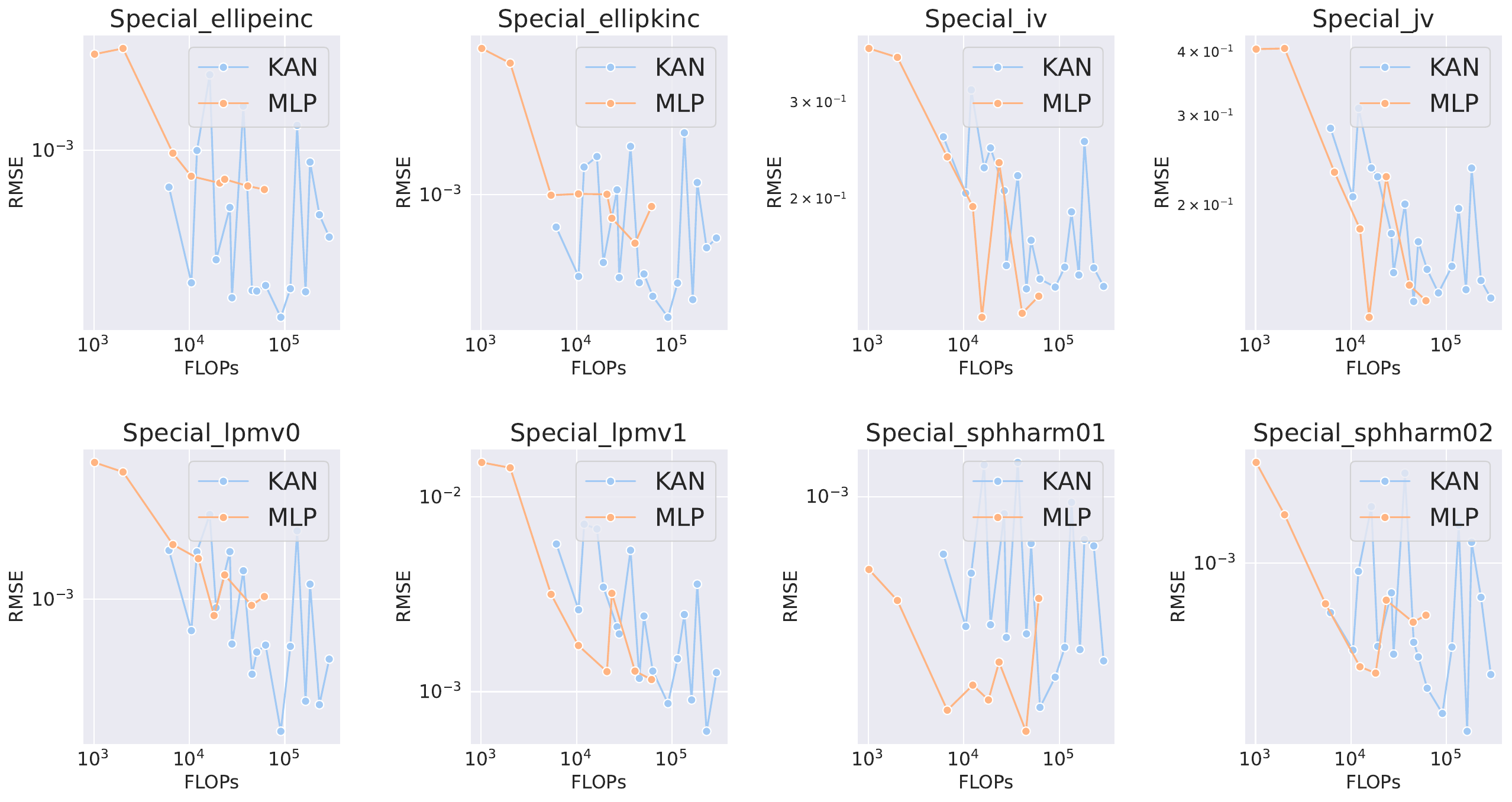}
  \caption{Performance comparison between KAN and MLP for symbolic formula representing controlling FLOPs.}
  \label{fig:special_flops}
\end{figure}

\textbf{Symbolic Formula Representing.} We compared the differences between KAN and MLP on 8 symbolic formula representing tasks. The functions to be fitted were sourced from Section 3.2 of the original KAN paper.  We use KAN and MLP with one to four hidden layers. The input and output dimensions of the neural networks were adjusted according to the datasets. The hidden layer widths for MLP were 5, 50, or 100, with GELU or ReLU as the activation functions, and no normalization layers were used in the MLP. For KAN, the hidden layer width was 5, the number of B-spline grids was 3, 5, 10, or 20, and the B-spline degrees were 2, 3, or 5. All experiments are trained for 500 epochs. The best RMSE on the testing set during the training is recorded. The lower envelope of all the recorded points is plotted in the \cref{fig:special_para} and \cref{fig:special_flops}. When controlling for the number of parameters, KAN outperformed MLP on 7 out of 8 datasets. When controlling for FLOPs, due to the additional computational complexity introduced by the spline functions, KAN's performance was roughly equivalent to MLP, outperforming MLP on two datasets and underperforming on one. Overall, for symbolic formula representation tasks, KAN performed better than MLP.

\begin{figure}[h!]
  \centering
  \includegraphics[width=0.9\textwidth]{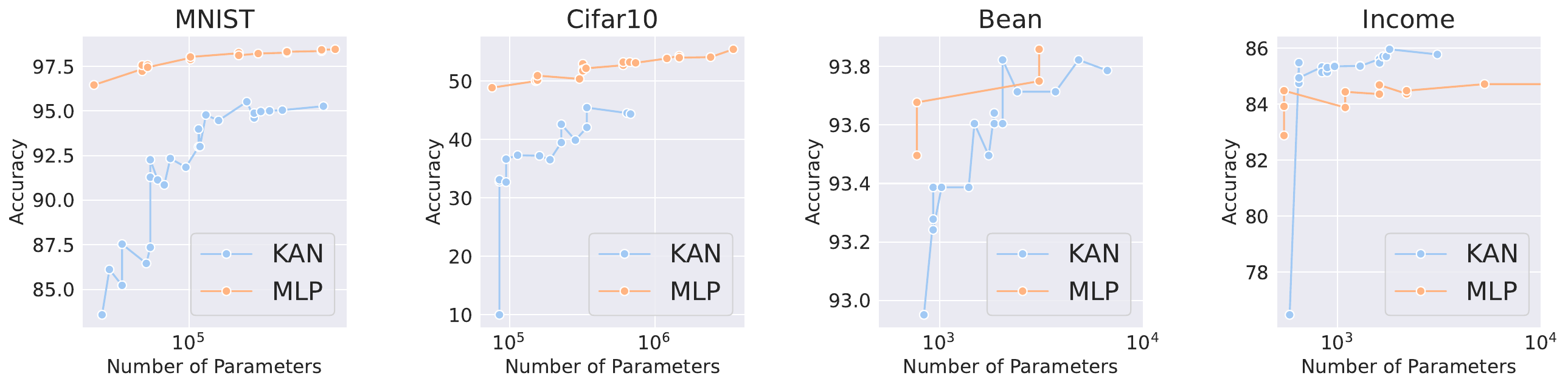}
  \caption{Performance comparison between KAN and MLP using LBFGS optimizer controlling the number of parameters.}
  \label{fig:lbfgs_para}
\end{figure}

\begin{figure}[h!]
  \centering
  \includegraphics[width=0.9\textwidth]{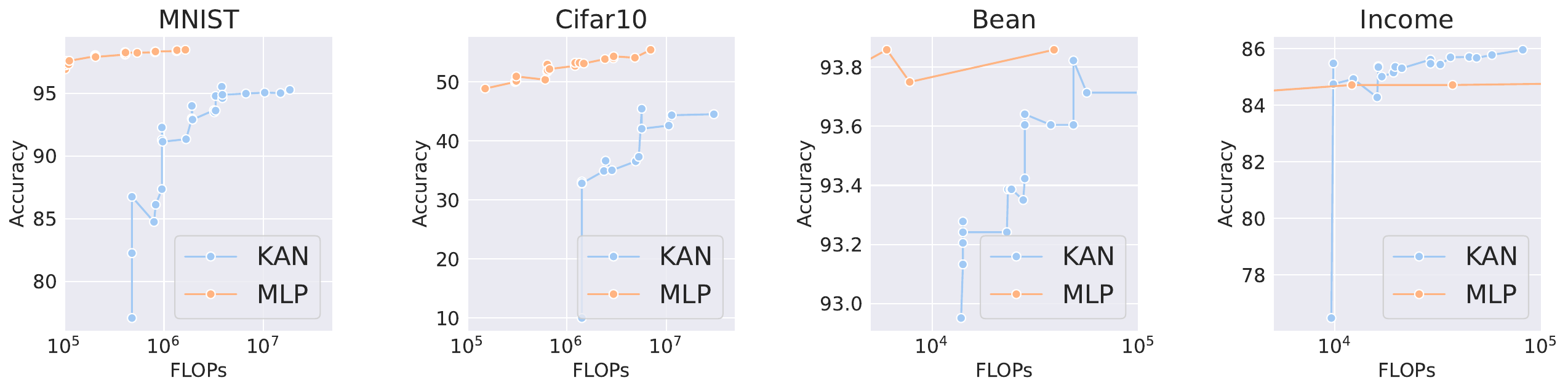}
  \caption{Performance comparison between KAN and MLP using LBFGS optimizer controlling FLOPs.}
  \label{fig:lbfgs_flops}
\end{figure}

\textbf{LBFGS Optimizer.} Instead of using the Adam optimizer as in the previous experiments, we used the LBFGS optimizer in the following experiments. The primary difference in implementation is that we increased the batch size to 1024 to improve the accuracy of the Hessian matrix approximation in LBFGS. The hidden layer numbers for both the MLP and KAN models are 1, 2, 3, or 5. The hidden layer widths for the MLP are 32, 64, and 128, while for KAN, they are 4, 8, 16, and 32. All other experimental settings remain consistent with the corresponding previous experiments. Results are plotted in \cref{fig:lbfgs_para,fig:lbfgs_flops}. The relative advantages between the KAN and MLP networks are largely the same as when using Adam. The MLP continues to have an advantage on the MNIST, CIFAR-10, and Bean datasets, while KAN maintains its advantage on the Income dataset.

\subsection{Architecture Ablation}
From the above experiments, we can conclude that KAN and MLP indeed have functional differences. KAN is more suitable for task symbolic formula representing, while MLP demonstrates advantages in machine learning, computer vision, NLP, and audio processing. Next, we perform ablation studies on the architecture of KAN and MLP to determine which specific components contribute to their differences.

\begin{figure}[h!]
  \centering
  \includegraphics[width=0.5\textwidth]{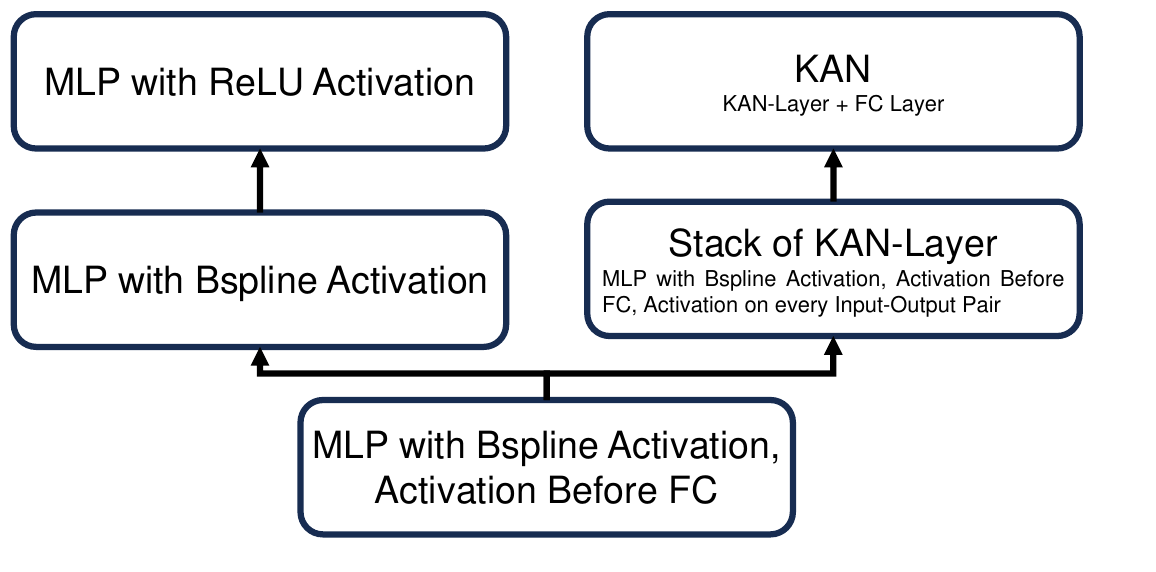}
  \caption{Architectures Included in the Ablation.}
  \label{fig:archs}
\end{figure}

We conducted ablation experiments on two computer vision datasets and two machine learning datasets. The architectures included in the ablation and their relationship are plotted in \cref{fig:archs}. KAN is a mix of KAN Layers and fully-connected Layers, thus the first architecture we ablate is built by purely stacking KAN Layers without fully-connected Layers. Because the KAN Layer is actually a fully-connected Layer with the special B-spline activation before the linear transformation. Thus, we also compare with MLP with B-spline activation before or after the linear transformation.

\begin{figure}[h!]
  \centering
  \includegraphics[width=0.9\textwidth]{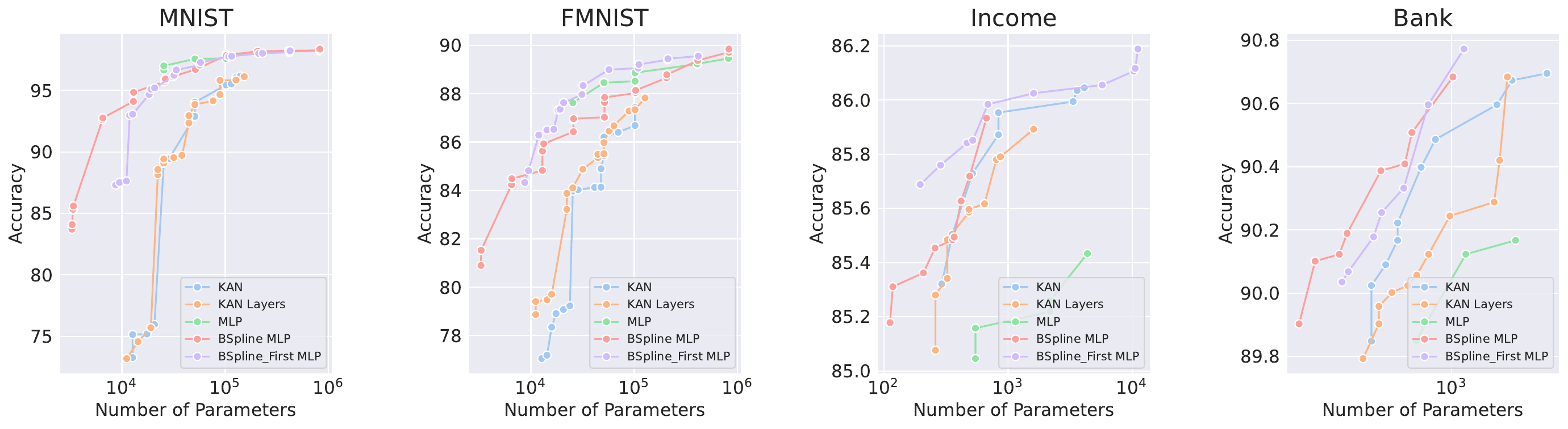}
  \caption{Architecture Ablation from MLP to KAN.}
  \label{fig:arch_abl}
\end{figure}

\begin{figure}[h!]
  \centering
  \includegraphics[width=0.9\textwidth]{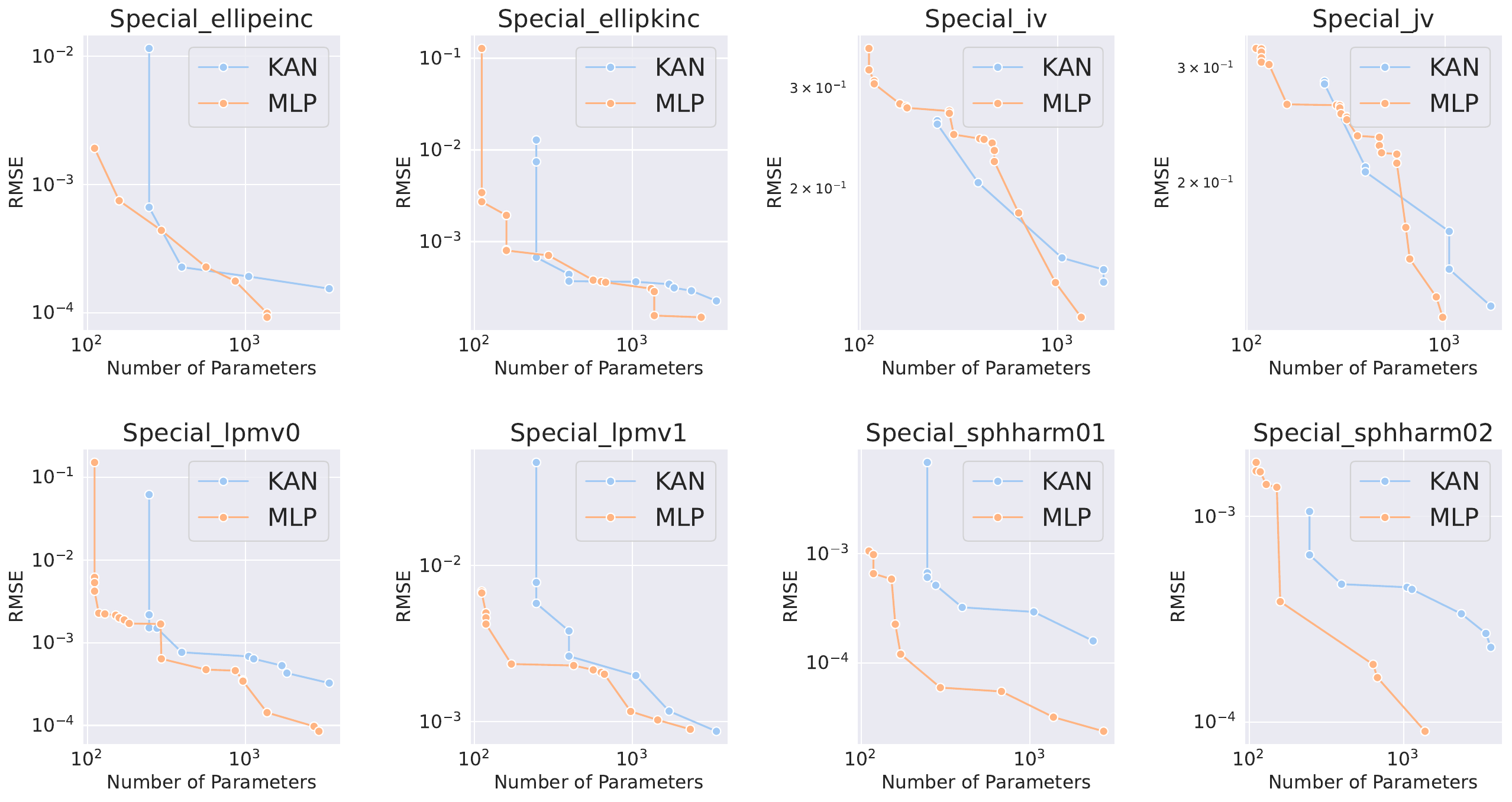}
  \caption{Performance comparison between KAN and MLP with spline activation for symbolic formula representing controlling the number of parameters.}
  \label{fig:arch_special_para}
\end{figure}

The observations are twofold. (1) On the computer vision datasets, for MLP models, using spline activation functions brought little improvement. However, the higher computational cost of spline functions makes it a worse choice than conventional ReLU or GELU activation functions. 
(2) However, on the other hand, for machine learning tasks, replacing the activation functions in MLP with spline functions resulted in a significant performance boost. 
Initially, on these two datasets, MLP's performance was inferior to KAN, but after substituting the activation function with a spline function, MLP's performance became comparable to KAN. This indicates that KAN's performance improvement on machine learning datasets primarily stems from the use of spline activation functions. Whether the linear transformation is applied before the spline function, or vice versa, does not significantly affect the final performance. Whether the spline activation is used on all input-output pairs or only on the inputs or outputs does not significantly affect the performance.

\begin{figure}[h!]
  \centering
  \includegraphics[width=0.9\textwidth]{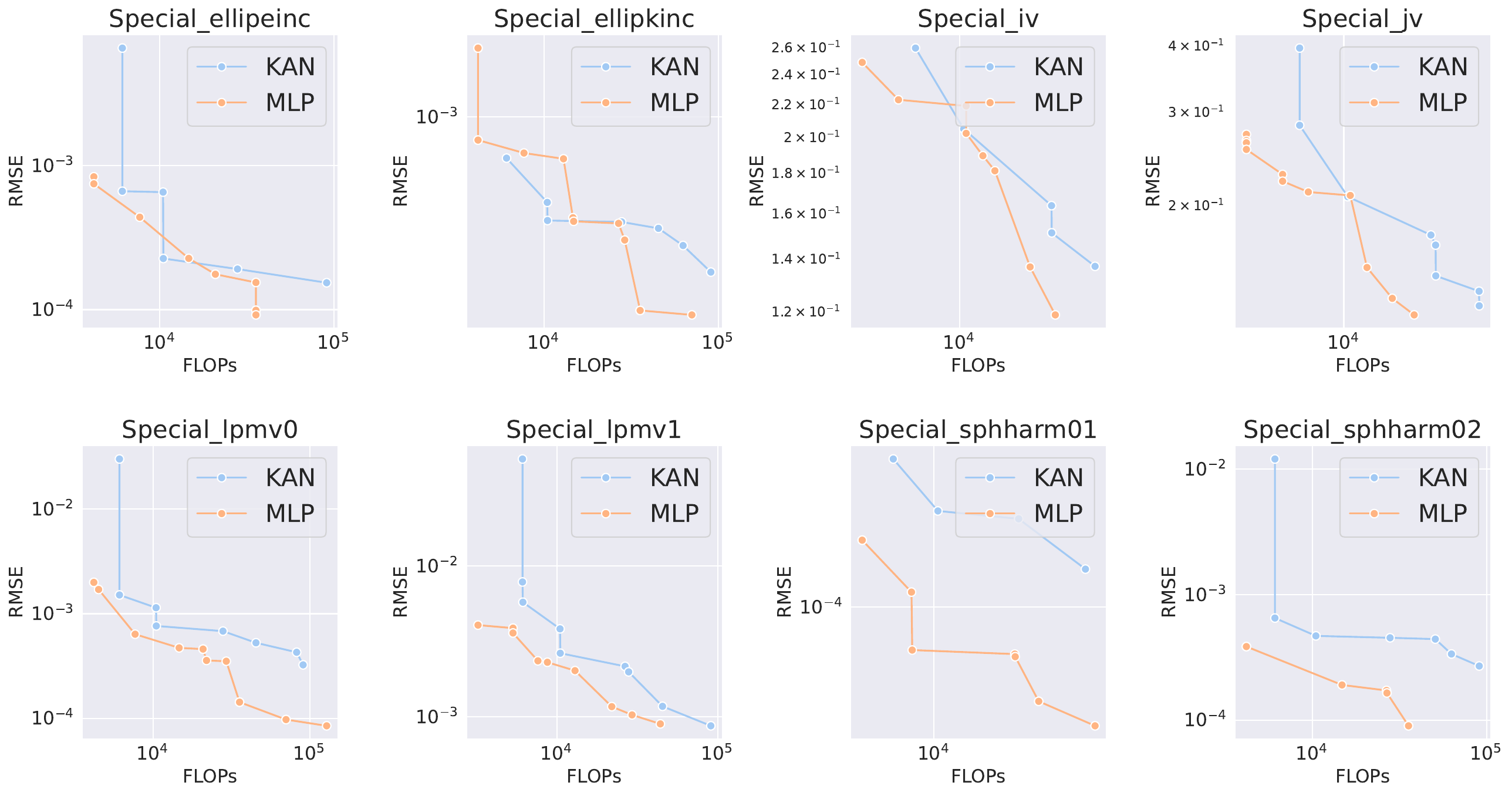}
  \caption{Performance comparison between KAN and MLP with Spline activation for symbolic formula representing controlling FLOPs.}
  \label{fig:arch_special_flops}
\end{figure}

We also evaluated the performance of MLP using B-spline activation functions on symbolic formula representing. In this experiment, the hidden layer width of MLP was set to 10 or 20 to balance the additional parameters and FLOPs introduced by the B-spline function. All other settings were the same as in the previous symbolic formula representing experiments. The results are shown in \cref{fig:arch_special_para} and \cref{fig:arch_special_flops}. In the previous symbolic formula representation experiments, KAN's performance is overall superior to that of MLP. However, with B-spline activation functions, MLP's performance in all symbolic formula representing tasks can either surpass or match that of KAN. This fully validates our hypothesis that the functional differences between KAN and MLP stem from the different activation functions they use. These different activation functions make the two architectures suited for different tasks.

\begin{table}[h!]
\centering
\footnotesize
\begin{tabular}{@{}c|c@{$\quad$}c@{$\quad$}c@{$\ $}c@{$\ $}c|c@{$\quad$}c@{$\quad$}c@{$\ $}c@{$\ $}c@{}}
\toprule
\multirow{2}{*}{\#Parameters} & \multicolumn{5}{c}{KAN}                                                                          & \multicolumn{5}{c}{MLP}                                                                          \\ \cmidrule(l){2-11} 
                              & $\overline{\text{ACC}}$ & Backward & T1 Acc & T2 Acc & T3 Acc & $\overline{\text{ACC}}$ & Backward & T1 Acc & T2 Acc & T3 Acc \\ \midrule
0.4M                          & 31.70            & -63.18            & 0.00              & 0.00              & 95.10             & 75.58            & -9.20             & 74.45             & 64.63             & 87.67             \\
0.3M                          & 29.82            & -59.15            & 0.00              & 0.00              & 89.46             & 72.84            & -13.00            & 65.40             & 65.92             & 87.20             \\
0.2M                          & 26.58            & -63.61            & 0.00              & 0.00              & 79.73             & 64.70            & -12.87            & 60.53             & 52.12             & 81.45             \\
0.1M                          & 20.39            & -50.39            & 0.00              & 0.00              & 61.18             & 62.93            & -5.67             & 58.21             & 57.73             & 72.84             \\ \midrule
\multirow{2}{*}{FLOPs}        & \multicolumn{5}{c}{KAN}                                                                          & \multicolumn{5}{c}{MLP}                                                                          \\ \cmidrule(l){2-11} 
                              & $\overline{\text{ACC}}$ & Backward & T1 Acc & T2 Acc & T3 Acc & $\overline{\text{ACC}}$ & Backward & T1 Acc & T2 Acc & T3 Acc \\ \midrule
3.7M                          & 31.96            & -63.76            & 0.00              & 0.00              & 95.88             & 66.98            & -29.02            & 50.49             & 55.03             & 95.41             \\
1.6M                          & 31.52            & -63.28            & 0.00              & 0.00              & 94.56             & 74.93            & -16.25            & 66.19             & 66.47             & 92.13             \\
1.3M                          & 29.90            & -60.09            & 0.00              & 0.00              & 89.70             & 78.06            & -13.79            & 70.58             & 72.50             & 91.11             \\
0.8M                          & 28.09            & -55.14            & 0.00              & 0.00              & 84.26             & 75.58            & -9.20             & 74.45             & 64.63             & 87.67             \\ \bottomrule
\end{tabular}
\vspace{1em}
\caption{Performance comparison between KAN and MLP under the continual learning setup.}\label{tab:continual_learning}
\end{table}

\subsection{Continual Learning}
The original paper of KAN validated that KAN performs better in the continual learning of one-dimensional functions. We further verified this in computer vision, by constructing a more standard class-incremental setup. We divided the MNIST dataset into three tasks according to digit: the first task involved classifying digits 0, 1, and 2; the second task involved classifying digits 3, 4, and 5; and the third task involved classifying digits 6, 7, 8, and 9. We trained KAN and MLP models on these three tasks sequentially. The training hyperparameters were the same as those used in the vision experiments, but the training epoch for each task was set to 1. The results are summarized in the \cref{tab:continual_learning}. It is evident that KAN did not exhibit an advantage in continual learning tasks for computer vision. Its speed of forgetting is faster than that of MLP. After the training of all three tasks, KAN's accuracy on the first and second tasks dropped to 0, achieving relatively high accuracy only on the third task. In contrast, MLP was able to retain acceptable accuracy on the first two tasks. The methods for calculating the average accuracy and backward scores are consistent with those in \cite{cl_survey}.

\section{Related Works}
The KAN was introduced in \cite{liu2024kan}, with related ideas also discussed in \cite{kan_idea_1,kan_idea_2}. 
Significant improvements have been made to KAN networks. In KAN, the nonlinear activation originally present in MLP is implemented using B-spline functions. Due to the locality and adjustable number of grids in B-splines, KAN networks can achieve a certain degree of dynamic network architecture and continual learning. Numerous studies have replaced B-splines with Chebyshev polynomials~\cite{ss2024chebyshev}, wavelet functions~\cite{bozorgasl2024wavkan}, Jacobi polynomials~\cite{jacobiKAN}, orthogonal polynomials~\cite{orthKAN}, etc., imparting different properties to KAN networks. More efforts have been made to combine KAN networks with existing popular network structures for various applications. For instance, integrating KAN with convolutional networks and vision transformer networks to enhance KAN's performance in classification and other vision tasks~\cite{VisionKAN2024,convKAN1,convKAN2,convKAN3,convKAN4,convKAN5,convKAN6,convKAN7}, combining KAN with U-Net for medical image segmentation and generation~\cite{li2024ukan}, merging KAN with GNN for graph-related tasks~\cite{li2024ukan}, or integrating KAN with NeRF networks for 3D reconstruction~\cite{XKANeRF}. 
Different from these works, this paper does not introduce any improvements over KAN. Instead, we provide a fairer and comprehensive comparison between KAN and MLP  to offer insights for future research. 

\section{Conclusion}
In this work, we first mathematically compared the forward process of KAN and MLP. We found that KAN can be seen as a special type of MLP, with its uniqueness stemming from the use of learnable B-spline functions as activation functions. We hypothesize that this difference in activation functions is the primary reason for the functional differences between KAN and MLP. To test this hypothesis, we compared the performance of KAN and MLP on symbolic formula representing, machine learning, computer vision, natural language processing, and audio processing tasks, controlling for the same number of parameters or FLOPs. Our experiments revealed that KAN only had an advantage in symbolic formula representing, while MLP outperformed KAN in other tasks. Furthermore, we found that after replacing MLP's activation function with the learnable B-spline, MLP outperformed or was comparable to KAN across all tasks. Lastly, we discovered that under standard class-incremental learning settings, KAN exhibited more severe forgetting issues compared to MLP.

\clearpage
{
    \small
    \bibliographystyle{ieeenat_fullname}
    \bibliography{main}
}

\end{document}